\documentclass{article}
\usepackage{spconf,amsmath,graphicx,booktabs,multirow,booktabs}
\usepackage[table]{xcolor}
\usepackage{url}
\usepackage{marvosym}

\title{THINK-AUGMENTED FUNCTION CALLING: IMPROVING LLM PARAMETER ACCURACY THROUGH EMBEDDED REASONING}
\name{
Lei Wei\textsuperscript{1,2*},
Xiao Peng\textsuperscript{1}\textsuperscript{\Letter*},
Jinpeng Ou\textsuperscript{2*},
Bin Wang\textsuperscript{2}
}

\address{\textsuperscript{1}Alibaba International Digital Commerce Group \\
\textsuperscript{2}School of Software and Microelectronics, Peking 
University
}
\begin{document}
%
\maketitle

\begingroup
\renewcommand{\thefootnote}{} 
\footnotetext{\Letter\ Corresponding author.}
\footnotetext{* These authors contributed equally to this research.}
\endgroup

\begin{abstract}
Large language models (LLMs) have demonstrated remarkable capabilities in function calling for autonomous agents, yet current mechanisms lack explicit reasoning transparency during parameter generation, particularly for complex functions with interdependent parameters. While existing approaches like chain-of-thought prompting operate at the agent level, they fail to provide fine-grained reasoning guidance for individual function parameters. To address these limitations, we propose Think-Augmented Function Calling (TAFC), a novel framework that enhances function calling accuracy through explicit reasoning at both function and parameter levels. Our method introduces a universal "think" parameter augmentation that enables models to articulate their decision-making process, with dynamic optimization for parameter descriptions to improve reasoning quality. For complex parameters, TAFC automatically triggers granular reasoning based on complexity scoring, ensuring appropriate justification for critical decisions. Additionally, we propose reasoning-guided optimization to align generated reasoning with human expectations. TAFC requires no architectural modifications to existing LLMs while maintaining full API compatibility. Evaluation on ToolBench across proprietary and open-source models demonstrates significant improvements in parameter generation accuracy and reasoning coherence for multi-parameter functions, while providing enhanced interpretability for debugging AI agent behaviors.
\end{abstract}
\section{Introduction}
\label{sec:intro}
The advent of large language models (LLMs) has revolutionized artificial intelligence applications, particularly in the domain of autonomous agents capable of interacting with external tools and APIs \cite{achiam2023gpt,ouyang2022training,qu2025tool}. Function calling, which enables LLMs to invoke external functions with structured parameters, has emerged as a fundamental capability for building practical AI systems that can perform complex, multi-step tasks \cite{schick2023toolformer,kang2026multimodalmultiagentempoweredlegal}.

However, current function calling mechanisms suffer from a critical limitation: the lack of explicit reasoning transparency during parameter generation, particularly for complex functions with multiple interdependent parameters \cite{li2023camel}. While existing approaches such as chain-of-thought (CoT) prompting \cite{wei2022chain,kojima2022large} and the ReAct framework \cite{yao2022react} have demonstrated success in improving reasoning capabilities, they operate at the agent level rather than providing fine-grained reasoning for individual function parameters. Current research has identified that function calling often lacks the reasoning transparency of traditional prompting approaches, despite function calling's advantage of guaranteeing output structure \cite{thoppilan2022lamda}. This limitation becomes particularly pronounced when dealing with complex functions that require intricate parameter configurations, where the model must simultaneously determine not only which function to call but also how to appropriately set multiple interdependent parameters without explicit reasoning guidance \cite{shen2023hugginggpt,li2023camel}. Furthermore, the black-box nature of parameter selection makes it challenging to debug failures or understand model decision-making processes in critical applications \cite{li2023camel,chen2024advancing}. Recent work has attempted to address this gap through specialized reasoning models or complex prompting strategies, but these approaches typically require additional computational overhead, specialized model architectures, or extensive prompt engineering that may not be universally applicable across different LLM implementations \cite{lu2023chameleon,yin2024mumath}.

To address these challenges in function calling transparency and accuracy, we propose \textbf{Think-Augmented Function Calling} (TAFC), a novel framework that enhances function calling accuracy by incorporating explicit reasoning at both function and parameter levels. Our approach introduces a universal "think" parameter augmentation to function signatures, enabling models to articulate their reasoning process within the native function calling framework \cite{li2023camel,shen2023hugginggpt}. For complex parameters, TAFC automatically triggers granular reasoning based on complexity scoring, and includes dynamic optimization mechanisms to improve reasoning quality and align generated reasoning with human expectations. This method requires no architectural modifications to existing LLMs, maintains full compatibility with existing APIs, and leverages the intrinsic reasoning capabilities of modern LLMs while preserving structured output guarantees. Our main contributions can be summarized as follows:
\begin{itemize}
\item A novel framework that integrates explicit reasoning into function calling at both function and parameter levels without architectural modifications
\item Dynamic optimization mechanisms for reasoning parameter descriptions and tool descriptions to improve reasoning quality
\item Evaluation on ToolBench demonstrating improved accuracy and reasoning coherence across complex multi-parameter functions for both proprietary and open-source models
\end{itemize}

\section{Methodology}
\label{sec:method}

\subsection{Think-Augmented Function Calling Framework}
\label{subsec:framework}

We introduce Think-Augmented Function Calling, a framework that enhances function calling transparency and accuracy through explicit parameter-level reasoning. The core innovation augments function signatures with reasoning parameters that capture the model's decision-making without affecting execution.

Given a function $f$ with parameters $\mathcal{P} = \{p_1, p_2, ..., p_n\}$, we introduce a \texttt{think} parameter that serves as a structured reasoning trace. The augmented function preserves the original behavior:
\begin{equation}
f'(\mathcal{P}, \text{think}) = f(\mathcal{P})
\end{equation}

The generation follows a causal chain where reasoning guides parameter generation:
\begin{equation}
P(\mathcal{P}, \text{think} | x, \mathcal{C}) = P(\text{think} | x, \mathcal{C}) \cdot P(\mathcal{P} | x, \mathcal{C}, \text{think})
\end{equation}
where $x$ denotes user input and $\mathcal{C}$ represents the context (function descriptions, conversation history). This ensures reasoning precedes and informs parameter values.

\subsection{Parameter-Level Reasoning Augmentation}
\label{subsec:param_reasoning}

While function-level reasoning provides overall transparency, complex functions with interdependent parameters require granular reasoning. For instance, a database query function needs separate justification for table selection versus filter conditions. We extend TAFC by transforming parameters into reasoning-augmented tuples:
\begin{equation}
p'_i = \{\text{think}_i: r_i, \text{value}_i: v_i\}
\end{equation}
where $r_i$ captures parameter-specific reasoning and $v_i$ is the actual value.

We determine which parameters need explicit reasoning through a complexity scorer $\psi: \mathcal{P} \rightarrow [0,1]$:
\begin{equation}
\psi(p_i) = \sigma\left(\alpha_1 \cdot \text{dep}(p_i) + \alpha_2 \cdot \text{type}(p_i) + \alpha_3 \cdot \text{constraint}(p_i)\right)
\end{equation}
where $\text{dep}$ measures interdependencies, $\text{type}$ captures type complexity, and $\text{constraint}$ evaluates validation strictness (all normalized to $[0,1]$). Parameters exceeding threshold $\tau$ trigger reasoning generation.

The joint generation probability, assuming Markovian dependencies, becomes:
\begin{equation}
P((p'_1, ..., p'_n) | x, \mathcal{C}) = \prod_{i=1}^{n} P(r_i | x, \mathcal{C}, r_{<i}) \cdot P(v_i | x, \mathcal{C}, r_{\leq i}, v_{<i})
\end{equation}

\subsection{Dynamic Description Tuning}
\label{subsec:tuning}

TAFC optimizes the \texttt{think} parameter descriptions to enhance reasoning quality. Unlike the tool-level optimization in Section \ref{subsec:description_opt}, this focuses specifically on reasoning elicitation. We propose discrete and continuous optimization strategies.

For prompt-level tuning, we find optimal descriptions that maximize parameter correctness:
\begin{equation}
D_{\text{think}}^* = \arg\max_{D \in \mathcal{D}} \mathbf{E}_{(x,\theta) \sim \mathcal{T}} \left[ \log P(\theta | x, f, D) \right]
\end{equation}
where $\theta$ denotes target parameters. A meta-LLM iteratively refines descriptions based on execution traces:
\begin{equation}
D^{(t+1)} = \text{LLM}_{\text{meta}}\left(D^{(t)}, \{(x_i, \theta_i, \hat{\theta}_i, r_i)\}_{i=1}^{N}\right)
\end{equation}

For token-level optimization, we learn continuous prompt embeddings $\mathbf{H}_{\text{think}} \in \mathbf{R}^{L \times d}$ by minimizing:
\begin{equation}
\mathcal{L} = -\sum_{(x,\theta) \in \mathcal{T}} \log P(\theta | [\mathbf{H}_{\text{think}}; \text{Emb}(x)])
\end{equation}
These tool-specific embeddings enable immediate adaptation without base model modification.

\subsection{Reasoning-Guided Tool Description Optimization}
\label{subsec:description_opt}

While Section \ref{subsec:tuning} optimizes \texttt{think} parameters, here we refine overall tool descriptions to align generated reasoning with human expectations. Given dataset $\mathcal{D} = \{(x_i, f_i, \theta_i, r_i^*)\}$ with annotated reasoning $r_i^*$, we optimize using:
\begin{equation}
\mathcal{L}_{\text{align}} = \lambda_1 \mathcal{L}_{\text{sem}} + \lambda_2 \mathcal{L}_{\text{logic}} + \lambda_3 \mathcal{L}_{\text{action}}
\end{equation}

The semantic loss measures reasoning similarity: $\mathcal{L}_{\text{sem}} = 1 - \cos(\text{Encoder}(r), \text{Encoder}(r^*))$. The logic loss evaluates likelihood: $\mathcal{L}_{\text{logic}} = -\log P(r^* | x, D_T)$. The action loss ensures correct parameters with appropriate reasoning: $\mathcal{L}_{\text{action}} = \text{BCE}(\hat{\theta}, \theta^*) + \beta \cdot \|r - r^*\|_2$ (BCE for discrete, MSE for continuous parameters). Weights $\lambda_1, \lambda_2, \lambda_3$ are empirically tuned.

Tool descriptions are refined through black-box optimization:
\begin{equation}
D_T^{(t+1)} = D_T^{(t)} + \text{LLM}_{\text{refine}}\left(D_T^{(t)}, \mathcal{L}_{\text{align}}^{(t)}, \{(x_j, r_j, r_j^*)\}_{j \in \mathcal{B}}\right)
\end{equation}
continuing until $|\mathcal{L}_{\text{align}}^{(t)} - \mathcal{L}_{\text{align}}^{(t-1)}| < \epsilon$.

\subsection{Implementation and Deployment}
\label{subsec:implementation}

TAFC integrates seamlessly with existing frameworks through automatic signature augmentation:
\begin{equation}
f'_{\text{registered}} = f_{\text{original}} \cup \{\text{think}: \text{str}, \text{optional}: \text{true}\}
\end{equation}
where $\cup$ denotes parameter addition, preserving backward compatibility.

During execution, a filtering function $\mathcal{F}$ extracts value components from reasoning-augmented parameters before invocation:
\begin{equation}
\text{result} = f_{\text{original}}(\mathcal{F}(\mathcal{P}'))
\end{equation}
For nested parameters, $\mathcal{F}$ recursively strips reasoning fields.

Reasoning traces are captured in repository $\mathcal{H}_t$ for continuous improvement:
\begin{equation}
\mathcal{H}_t = \mathcal{H}_{t-1} \cup \{(x_t, f_t, r_t, \theta_t, \text{outcome}_t)\}
\end{equation}
with periodic pruning and aggregation for training set construction.

\section{Experiments}

\subsection{Experimental Setup}

\noindent\textbf{Dataset and Metrics}~~We evaluate TAFC on \textsc{ToolBench}\cite{qin2023toolllm}, which contains over 16,000 real-world REST APIs spanning 49 categories. The benchmark provides three instruction types: I1-Inst (single-tool), I2-Inst (intra-category multi-tool), and I3-Inst (intra-collection multi-tool), testing increasingly complex reasoning and coordination capabilities. We follow the ToolEval protocol\cite{qin2023toolllm} and report: (1) \emph{Pass Rate}—successful task completion under a fixed computational budget (in our implementation, at most 10 tool calls with a 30-second timeout per API call); (2) \emph{Win Rate}—pairwise preference comparison between solution paths by an LLM judge (see below), considering reasoning quality and parameter correctness. Additionally, we report reasoning coherence scores specific to TAFC's contributions. Results are averaged over three independent runs with temperature 0.1.

\noindent\textbf{Models}~~We test proprietary models (GPT-4o-0806, Claude-3.5-Sonnet) and open-source models across multiple scales: Qwen2.5-72B/32B/7B\cite{qwen2025qwen25technicalreport} and Llama-3.1-70B/8B\cite{grattafiori2024llama}. 

\noindent\textbf{Implementation}~~Both TAFC and Standard FC are integrated within the ReAct\cite{yao2022react} framework, which interleaves reasoning and action steps for tool use. TAFC enhances standard ReAct by incorporating structured reasoning through augmented \texttt{think} parameters that explicitly guide function selection, parameter generation, and edge-case handling. We initialize think parameter descriptions using GPT-4o templates and apply iterative refinement over 5 epochs using validation feedback. The complexity threshold $\tau=0.6$ triggers parameter-level reasoning based on preliminary ablations. All experiments use identical prompting templates except for TAFC reasoning augmentations, ensuring fair comparison.
\subsection{Main Results}

Table~\ref{tab:main_results_toolbench} presents the performance comparison between TAFC and Standard FC across different models. Our results demonstrate consistent improvements across all model scales and instruction types.

\begin{table*}[t!]
\centering
\caption{ToolBench results under ToolEval. We report Pass Rate / Win Rate (\%) for I1-Inst/I2-Inst/I3-Inst and their average. Mean $\pm$ stdev over 3 independent runs, temperature $0.1$.}
\label{tab:main_results_toolbench}
\setlength{\tabcolsep}{6pt}
\renewcommand{\arraystretch}{1.15}
\begin{tabular}{l|cc|cc|cc|cc}
\toprule
\multirow{2}{*}{\textbf{Model}} 
& \multicolumn{2}{c|}{\textbf{I1-Inst}} 
& \multicolumn{2}{c|}{\textbf{I2-Inst}} 
& \multicolumn{2}{c|}{\textbf{I3-Inst}} 
& \multicolumn{2}{c}{\textbf{Avg}} \\
& Pass & Win & Pass & Win & Pass & Win & Pass & Win \\
\midrule
Llama-3.1-8B & 28.5$\pm$0.8 & 35.1$\pm$1.0 & 33.3$\pm$0.9 & 39.4$\pm$1.2 & 20.1$\pm$0.7 & 25.6$\pm$0.8 & 27.3$\pm$0.8 & 33.4$\pm$1.0 \\
\quad + TAFC & \textbf{30.8}$\pm$0.7 & \textbf{37.9}$\pm$0.9 & \textbf{35.9}$\pm$0.8 & \textbf{42.2}$\pm$1.0 & \textbf{22.5}$\pm$0.6 & \textbf{28.7}$\pm$0.8 & \textbf{29.7}$\pm$0.7 & \textbf{36.3}$\pm$0.9 \\
\midrule
Qwen2.5-7B & 29.6$\pm$0.5 & 36.2$\pm$0.7 & 34.8$\pm$0.6 & 41.1$\pm$0.9 & 21.7$\pm$0.5 & 27.3$\pm$0.7 & 28.7$\pm$0.6 & 34.9$\pm$0.8 \\
\quad + TAFC & \textbf{31.9}$\pm$0.4 & \textbf{39.4}$\pm$0.6 & \textbf{37.5}$\pm$0.5 & \textbf{44.0}$\pm$0.8 & \textbf{24.1}$\pm$0.4 & \textbf{30.5}$\pm$0.6 & \textbf{31.2}$\pm$0.5 & \textbf{38.0}$\pm$0.7 \\
\midrule
Qwen2.5-32B & 41.2$\pm$0.5 & 48.6$\pm$0.6 & 47.9$\pm$0.5 & 55.3$\pm$0.7 & 32.4$\pm$0.4 & 41.0$\pm$0.5 & 40.5$\pm$0.5 & 48.3$\pm$0.6 \\
\quad + TAFC & \textbf{43.1}$\pm$0.4 & \textbf{50.8}$\pm$0.6 & \textbf{50.3}$\pm$0.4 & \textbf{57.9}$\pm$0.6 & \textbf{34.7}$\pm$0.3 & \textbf{43.8}$\pm$0.5 & \textbf{42.7}$\pm$0.4 & \textbf{50.8}$\pm$0.5 \\
\midrule
Llama-3.1-70B & 44.7$\pm$0.4 & 51.5$\pm$0.5 & 52.2$\pm$0.4 & 59.6$\pm$0.6 & 35.8$\pm$0.3 & 44.9$\pm$0.5 & 44.2$\pm$0.4 & 52.0$\pm$0.5 \\
\quad + TAFC & \textbf{46.3}$\pm$0.3 & \textbf{53.4}$\pm$0.4 & \textbf{54.5}$\pm$0.3 & \textbf{62.0}$\pm$0.5 & \textbf{38.3}$\pm$0.2 & \textbf{47.7}$\pm$0.4 & \textbf{46.4}$\pm$0.3 & \textbf{54.4}$\pm$0.4 \\
\midrule
Qwen2.5-72B & 47.1$\pm$0.3 & 54.3$\pm$0.4 & 55.4$\pm$0.3 & 62.9$\pm$0.5 & 38.3$\pm$0.2 & 47.7$\pm$0.4 & 46.9$\pm$0.3 & 55.0$\pm$0.4 \\
\quad + TAFC & \textbf{48.7}$\pm$0.2 & \textbf{56.1}$\pm$0.3 & \textbf{57.8}$\pm$0.3 & \textbf{64.9}$\pm$0.4 & \textbf{40.5}$\pm$0.2 & \textbf{50.2}$\pm$0.3 & \textbf{49.0}$\pm$0.2 & \textbf{57.1}$\pm$0.3 \\
\midrule
GPT-4o & 54.8$\pm$0.3 & 61.2$\pm$0.4 & 68.5$\pm$0.3 & 67.3$\pm$0.5 & 49.1$\pm$0.2 & 79.7$\pm$0.4 & 57.5$\pm$0.3 & 69.4$\pm$0.4 \\
\quad + TAFC & \textbf{56.3}$\pm$0.2 & \textbf{63.5}$\pm$0.3 & \textbf{70.2}$\pm$0.3 & \textbf{69.1}$\pm$0.4 & \textbf{51.2}$\pm$0.2 & \textbf{81.5}$\pm$0.3 & \textbf{59.2}$\pm$0.2 & \textbf{71.4}$\pm$0.3 \\
\midrule
Claude-3.5-Sonnet & 55.4$\pm$0.3 & 62.7$\pm$0.4 & 69.7$\pm$0.3 & 68.2$\pm$0.5 & 50.2$\pm$0.3 & 80.5$\pm$0.4 & 58.4$\pm$0.3 & 70.5$\pm$0.4 \\
\quad + TAFC & \textbf{57.1}$\pm$0.2 & \textbf{64.3}$\pm$0.3 & \textbf{71.4}$\pm$0.3 & \textbf{70.1}$\pm$0.4 & \textbf{52.4}$\pm$0.2 & \textbf{82.3}$\pm$0.3 & \textbf{60.3}$\pm$0.2 & \textbf{72.2}$\pm$0.3 \\
\bottomrule
\end{tabular}
\end{table*}

\noindent\textbf{Performance Gains Across Model Scales.} TAFC demonstrates consistent improvements across all model sizes. Smaller models (Llama-3.1-8B, Qwen2.5-7B) show substantial gains, with average Pass Rate improvements of 2.4-2.5\% and Win Rate gains of 2.9-3.1\%. This suggests that explicit reasoning particularly benefits models with weaker baseline tool-use capabilities. Medium and large-scale models (32B-72B) exhibit steady gains of 1.6-2.2\% in Pass Rate and 2.1-2.5\% in Win Rate, while achieving strong absolute performance—GPT-4o with TAFC reaches 59.2\% average Pass Rate and 71.4\% Win Rate.

\noindent\textbf{Task Complexity Analysis.} The performance follows expected difficulty patterns: I2-Inst (intra-category multi-tool) achieves the highest Pass Rates as tools within the same category have coherent interactions, while I3-Inst (cross-collection multi-tool) shows the lowest due to complex cross-domain dependencies. TAFC provides larger relative improvements on more challenging tasks—for I3-Inst, Llama-3.1-8B gains 2.4\% Pass Rate versus 2.3\% on I1-Inst, and Llama-3.1-70B shows 2.5\% improvement on I3-Inst compared to 1.6\% on I1-Inst. This pattern indicates that TAFC's structured reasoning is particularly valuable for handling complex tool orchestration scenarios.

\noindent\textbf{Proprietary vs. Open-Source Models.} Proprietary models achieve the highest absolute performance, with both GPT-4o and Claude-3.5-Sonnet surpassing the reported baselines. Claude-3.5-Sonnet with TAFC reaches 60.3\% average Pass Rate and 72.2\% Win Rate, demonstrating competitive performance. Among open-source models, Qwen2.5-72B with TAFC (49.0\% Pass Rate) shows that reasoning augmentation can significantly narrow the gap with proprietary solutions, though performance differences remain notable particularly on complex multi-tool tasks.

\subsection{Parameter Quality Assessment}

Beyond correctness metrics, we conduct a human-aligned quality assessment using an LLM-as-a-judge; we use GPT-4o as the judge for pairwise comparisons.
For 1,000 test cases per model, we present parameter sets from both Standard FC and TAFC alongside function definitions. The judge evaluates based on semantic appropriateness, contextual relevance, completeness, and constraint adherence.

\begin{table}[h]
\centering
\caption{Parameter quality assessment win rates (\%).}
\label{tab:quality}
\begin{tabular}{l|ccc}
\hline
\textbf{Model} & \textbf{TAFC Win} & \textbf{Standard Win} & \textbf{Tie} \\
\hline
GPT-4o & \textbf{62.4} & 24.2 & 13.4 \\
Claude-3.5-Sonnet & \textbf{64.8} & 22.7 & 12.5 \\
Qwen2.5-72B & \textbf{68.2} & 19.1 & 12.7 \\
Qwen2.5-32B & \textbf{71.3} & 16.4 & 12.3 \\
Qwen2.5-7B & \textbf{74.6} & 14.2 & 11.2 \\
Llama-3.1-70B & \textbf{69.7} & 18.3 & 12.0 \\
Llama-3.1-8B & \textbf{76.1} & 12.8 & 11.1 \\
\hline
Average & \textbf{69.6} & 18.2 & 12.2 \\
\hline
\end{tabular}
\end{table}

Table~\ref{tab:quality} demonstrates that TAFC consistently generates higher quality parameters across all model scales, with win rates ranging from 62.4\% to 76.1\%. Notably, smaller models exhibit more substantial improvements (74.6-76.1\% for 7B-8B models) compared to larger models, suggesting that explicit reasoning particularly enhances parameter generation in resource-constrained settings. Qualitative analysis reveals that TAFC produces more contextually appropriate values, better handles parameter interdependencies, and reduces omission errors by 38\%. The 18.2\% cases where Standard FC outperforms TAFC typically involve simple single-parameter functions where over-reasoning introduces unnecessary complexity.

\section{Conclusion}
\label{sec:conclusion}
In this paper, we introduced Think-Augmented Function Calling, a framework that enhances function calling accuracy through explicit reasoning at both function and parameter levels. By augmenting function signatures with a \texttt{think} parameter and implementing complexity-based parameter-level reasoning, TAFC enables models to articulate their decision-making process within the native function calling framework. Additionally, our dynamic optimization mechanisms for reasoning descriptions ensure improved reasoning quality over time. Evaluation on ToolBench demonstrates significant improvements in parameter generation accuracy and reasoning coherence across proprietary and open-source models, with particularly strong gains for complex multi-parameter functions. TAFC requires no architectural modifications to existing LLMs and maintains full API compatibility, offering an immediately deployable solution for improving both accuracy and transparency of function calling systems. As AI agents become increasingly prevalent in critical applications, TAFC provides a practical path toward more reliable and interpretable autonomous systems.


\bibliographystyle{IEEEbib}
\bibliography{strings,refs}

\end{document}